# Mining the Web for Synonyms:
# PMI-IR versus LSA on TOEFL

Peter D. Turney

Institute for Information Technology, National Research Council of Canada,
M-50 Montreal Road, Ottawa, Ontario, Canada, K1A 0R6
`peter.turney@nrc.ca`

**Abstract.** This paper presents a simple unsupervised learning algorithm for recognizing synonyms, based on statistical data acquired by querying a Web search engine. The algorithm, called PMI-IR, uses Pointwise Mutual Information (PMI) and Information Retrieval (IR) to measure the similarity of pairs of words. PMI-IR is empirically evaluated using 80 synonym test questions from the Test of English as a Foreign Language (TOEFL) and 50 synonym test questions from a collection of tests for students of English as a Second Language (ESL). On both tests, the algorithm obtains a score of 74%. PMI-IR is contrasted with Latent Semantic Analysis (LSA), which achieves a score of 64% on the same 80 TOEFL questions. The paper discusses potential applications of the new unsupervised learning algorithm and some implications of the results for LSA and LSI (Latent Semantic Indexing).

## 1    Introduction

This paper introduces a simple unsupervised learning algorithm for recognizing synonyms. The task of recognizing synonyms is, given a problem word and a set of alternative words, choose the member from the set of alternative words that is most similar in meaning to the problem word. The unsupervised learning algorithm performs this task by issuing queries to a search engine and analyzing the replies to the queries. The algorithm, called PMI-IR, uses Pointwise Mutual Information (PMI) [1, 2] to analyze statistical data collected by Information Retrieval (IR). The quality of the algorithm's performance depends on the size of the document collection that is indexed by the search engine and the expressive power of the search engine's query language. The results presented here are based on queries to the AltaVista search engine [3].

Recognizing synonyms is often used as a test to measure a (human) student's mastery of a language. I evaluate the performance of PMI-IR using 80 synonym test questions from the Test of English as a Foreign Language (TOEFL) [4] and 50 synonym test questions from a collection of tests for students of English as a Second Language (ESL) [5]. PMI-IR obtains a score of 73.75% on the 80 TOEFL questions (59/80) and 74% on the 50 ESL questions (37/50). By comparison, the average score on the 80 TOEFL questions, for a large sample of applicants to US colleges from non-English speaking countries, was 64.5% (51.6/80) [6]. Landauer and Dumais [6] note that, "…we have been told that the average score is adequate for admission to many universities."

Latent Semantic Analysis (LSA) is another unsupervised learning algorithm that has been applied to the task of recognizing synonyms. LSA achieves a score of 64.4% (51.5/80) on the 80 TOEFL questions [6]. Landauer and Dumais [6] write, regarding this score for LSA, "We know of no other fully automatic application of a knowledge acquisition and representation model, one that does not depend on knowledge being entered by a human but only on its acquisition from the kinds of experience on which a human relies, that has been capable of performing well on a full scale test used for adults." It is interesting that PMI-IR, which is conceptually simpler than LSA, scores almost 10% higher on the TOEFL questions.

LSA is a statistical algorithm based on Singular Value Decomposition (SVD). A variation on this algorithm has been applied to information retrieval, where it is known as Latent Semantic Indexing (LSI) [7]. The performance of LSA on the TOEFL test has been widely cited as evidence for the value of LSA and (by relation) LSI. In this paper, I discuss the implications of the new unsupervised learning algorithm and the synonym test results for LSA and LSI.

In the next section, I describe the PMI-IR algorithm. I then discuss related work on synonym recognition in Section 3. I briefly explain LSA in the following section. The experiments with the TOEFL questions and the ESL questions are presented in Sections 5 and 6, respectively. In Section 7, I discuss the interpretation of the results and their significance for LSA and LSI. The next section discusses potential applications of PMI-IR and the final section gives the conclusions.

## 2    PMI-IR

Consider the following synonym test question, one of the 80 TOEFL questions. Given the problem word *levied* and the four alternative words *imposed, believed, requested, correlated*, which of the alternatives is most similar in meaning to the problem word [8]? Let *problem* represent the problem word and {*choice₁, choice₂, …, choiceₙ*} represent the alternatives. The PMI-IR algorithm assigns a score to each choice, score(*choiceᵢ*), and selects the choice that maximizes the score.

The PMI-IR algorithm, like LSA, is based on co-occurrence [9]. The core idea is that "a word is characterized by the company it keeps" [10]. There are many different measures of the degree to which two words co-occur [9]. PMI-IR uses Pointwise Mutual Information (PMI) [1, 2], as follows:

$$\text{score}(choice_i) = \log_2(\text{p}(problem \ \& \ choice_i) \ / \ (\text{p}(problem)\text{p}(choice_i))) \tag{1}$$

Here, p(*problem & choiceᵢ*) is the probability that *problem* and *choiceᵢ* co-occur. If *problem* and *choiceᵢ* are statistically independent, then the probability that they co-occur is given by the product p(*problem*)p(*choiceᵢ*). If they are not independent, and they have a tendency to co-occur, then p(*problem & choiceᵢ*) will be greater than

p($problem$)p($choice_i$). Therefore the ratio between p($problem$ & $choice_i$) and p($problem$)p($choice_i$) is a measure of the degree of statistical dependence between $problem$ and $choice_i$. The log of this ratio is the amount of information that we acquire about the presence of $problem$ when we observe $choice_i$. Since the equation is symmetrical, it is also the amount of information that we acquire about the presence of $choice_i$ when we observe $problem$, which explains the term *mutual information*.[1]

Since we are looking for the maximum score, we can drop $\log_2$ (because it is monotonically increasing) and p($problem$) (because it has the same value for all choices, for a given problem word). Thus (1) simplifies to:

$$\text{score}(choice_i) = \text{p}(problem \text{ \& } choice_i) \text{ / p}(choice_i) \qquad \textbf{(2)}$$

In other words, each choice is simply scored by the conditional probability of the problem word, given the choice word, p($problem$ | $choice_i$).

PMI-IR uses Information Retrieval (IR) to calculate the probabilities in (2). In this paper, I evaluate four different versions of PMI-IR, using four different kinds of queries. The following description of these four different methods for calculating (2) uses the AltaVista Advanced Search query syntax [11]. Let hits($query$) be the number of hits (the number of documents retrieved) when the query $query$ is given to AltaVista. The four scores are presented in order of increasing sophistication. They can be seen as increasingly refined interpretations of what it means for two words to co-occur, or increasingly refined interpretations of equation (2).

**Score 1:** In the simplest case, we say that two words co-occur when they appear in the same document:

$$\text{score}_1(choice_i) = \text{hits}(problem \text{ AND } choice_i) \text{ / hits}(choice_i) \qquad \textbf{(3)}$$

We ask AltaVista how many documents contain both $problem$ and $choice_i$, and then we ask how many documents contain $choice_i$ alone. The ratio of these two numbers is the score for $choice_i$.

**Score 2:** Instead of asking how many documents contain both $problem$ and $choice_i$, we can ask how many documents contain the two words close together:

$$\text{score}_2(choice_i) = \text{hits}(problem \text{ NEAR } choice_i) \text{ / hits}(choice_i) \qquad \textbf{(4)}$$

The AltaVista NEAR operator constrains the search to documents that contain $problem$ and $choice_i$ within ten words of one another, in either order.

**Score 3:** The first two scores tend to score antonyms as highly as synonyms. For example, *big* and *small* may get the same score as *big* and *large*. The following score tends to reduce this effect, resulting in lower scores for antonyms:

---

[1] For an explanation of the term *pointwise* mutual information, see [9].

$$\text{score}_3(choice_i) =$$

$$\frac{\text{hits}((problem \text{ NEAR } choice_i) \text{ AND NOT } ((problem \text{ OR } choice_i) \text{ NEAR "not"}))}{\text{hits}(choice_i \text{ AND NOT } (choice_i \text{ NEAR "not"}))} \quad \textbf{(5)}$$

**Score 4:** The fourth score takes context into account. There is no context for the TOEFL questions, but the ESL questions involve context. For example [5], "Every year in the early spring farmers [tap] maple syrup from their trees (drain; boil; knock; rap)." The problem word *tap*, out of context, might seem to best match the choice words *knock* or *rap*, but the context *maple syrup* makes *drain* a better match for *tap*. In general, in addition to the problem word *problem* and the alternatives {$choice_1$, $choice_2$, …, $choice_n$}, we may have context words {$context_1$, $context_2$, …, $context_m$}. The following score includes a context word:

$$\text{score}_4(choice_i) =$$

$$\frac{\text{hits}((problem \text{ NEAR } choice_i) \text{ AND } context \text{ AND NOT } ((problem \text{ OR } choice_i) \text{ NEAR "not"}))}{\text{hits}(choice_i \text{ AND } context \text{ AND NOT } (choice_i \text{ NEAR "not"}))} \quad \textbf{(6)}$$

This equation easily generalizes to multiple context words, using AND, but each additional context word narrows the sample size, which might make the score more sensitive to noise (and could also reduce the sample size to zero). To address this issue, I chose only one context word from each ESL question. For a given ESL question, I automatically selected the context word by first eliminating the problem word (*tap*), the alternatives (*drain, boil, knock, rap*), and stop words (*in, the, from, their*). The remaining words (*every, year, early, spring, farmers, maple, syrup, trees*) were context words. I then used p(*problem* | *context_i*), as calculated by score$_3$(*context_i*), to evaluate each context word. In this example, *syrup* had the highest score (*maple* was second highest; that is, *maple* and *syrup* have the highest semantic similarity to *tap*, according to score$_3$), so *syrup* was selected as the context word *context* for calculating score$_4$(*choice_i*).

## 3    Related Work

There are several well-known lexical database systems that include synonym information, such as WordNet [12], BRICO [13], and EuroWordNet [14]. These systems were constructed by hand, without machine learning, which ensures a certain level of quality, at the cost of a substantial amount of human labour. A major limitation of such hand-generated lexicons is the relatively poor coverage of technical and scientific terms. For example, I am interested in applying synonym recognition algorithms to the automatic extraction of keywords from documents [15]. In a large collection of scientific and technical journals, I found that only about 70% of the authors' keywords were in WordNet. (On the other hand, 100% were indexed by AltaVista.) This is a strong motivation for automating aspects of the construction of lexical databases. Another motivation is that the labour involved must be repeated for each new language and must be repeated regularly as new terms are added to a language.

Statistical approaches to synonym recognition are based on co-occurrence [9]. Manning and Schütze distinguish between co-occurrence (or association) and collocation: *collocation* refers to "grammatically bound elements that occur in a particular order", but *co-occurrence* and *association* refer to "the more general phenomenon of words that are likely to be used in the same context" [9]. Order does not matter for synonyms, so we say that they co-occur, rather than saying that they are collocated. Pointwise Mutual Information (PMI) has primarily been applied to analysis of collocation, but there have been some applications to co-occurrence analysis [1, 2]. I believe that the novelty in PMI-IR is mainly the observation that PMI can exploit IR. Instead of analyzing a document collection from scratch, specifically for co-occurrence information, we can take advantage of the huge document collections that have been indexed by modern Web search engines.

Various measures of semantic similarity between word pairs have been proposed, some using statistical (unsupervised learning from text) techniques [16, 17, 18], some using lexical databases (hand-built) [19, 20], and some hybrid approaches, combining statistics and lexical information [21, 22]. Statistical techniques typically suffer from the *sparse data problem*: they perform poorly when the words are relatively rare, due to the scarcity of data. Hybrid approaches attempt to address this problem by supplementing sparse data with information from a lexical database [21, 22]. PMI-IR addresses the sparse data problem by using a huge data source: the Web. As far as I know, no previous work in the statistical approach to semantic similarity has been able to exploit such a large body of text.

Another popular statistical approach to measuring semantic similarity is Latent Semantic Analysis (LSA) [6, 7, 8]. I will discuss this approach in the next section.

The work described in this paper is also related to the literature on data mining and text mining, in that it presents a method for extracting interesting relational information from a very large database (AltaVista). The most closely related work is the use of *interest* to discover interesting associations in large databases [23]. The *interest* of an association $A$ & $B$ is defined as p($A$ & $B$) / (p($A$)p($B$)). This is clearly equivalent to PMI without the log function (see equation (1) above). As far as I know, *interest* has been applied to data mining, but not to text mining.

## 4      Latent Semantic Analysis

LSA uses the Singular Value Decomposition (SVD) to analyze the statistical relationships among words in a collection of text [6, 7, 8]. The first step is to use the text to construct a matrix $\mathbf{X}$, in which the row vectors represent words and the column vectors represent chunks of text (e.g., sentences, paragraphs, documents). Each cell represents the *weight* of the corresponding word in the corresponding chunk of text. The *weight* is typically the TF.IDF score (Term Frequency times Inverse Document Frequency) for the word in the chunk. (TF.IDF is a standard tool in Information Retrieval.) The next step is to apply SVD to $\mathbf{X}$, to decompose $\mathbf{X}$ into a product of three matrices $\mathbf{ULA}^T$, where $\mathbf{U}$ and $\mathbf{A}$ are in column orthonormal form (i.e., the columns are orthogonal and have unit length) and $\mathbf{L}$ is a diagonal matrix of *singular values* (hence SVD). If $\mathbf{X}$ is of rank $r$, then $\mathbf{L}$ is also of rank $r$. Let $\mathbf{L}_k$, where $k < r$, be the matrix produced by removing from $\mathbf{L}$ the $r$ - $k$ columns and rows with the smallest singular values, and let $\mathbf{U}_k$ and $\mathbf{A}_k$ be the matrices produced by removing the corresponding columns from $\mathbf{U}$ and $\mathbf{A}$. The matrix $\mathbf{U}_k\mathbf{L}_k\mathbf{A}_k^T$ is the matrix of rank $k$ that best approximates the original matrix $\mathbf{X}$, in the sense that it minimizes the sum of the squares of the approximation errors. We may think of this matrix $\mathbf{U}_k\mathbf{L}_k\mathbf{A}_k^T$ as a "smoothed" or "compressed" version of the original matrix $\mathbf{X}$. SVD may be viewed as a form of *principal components* analysis. LSA works by measuring the similarity of words using this compressed matrix, instead of the original matrix. The similarity of two words is measured by the cosine of the angle between their corresponding compressed row vectors.

When they applied LSA to the TOEFL questions, Landauer and Dumais used an encyclopedia as the text source, to build a matrix $\mathbf{X}$ with 61,000 rows (words) and 30,473 columns (chunks of text; each chunk was one article from the encyclopedia) [6, 8]. They used SVD to generate a reduced matrix of rank 300. When they measured the similarity of the words (row vectors) in the original matrix $\mathbf{X}$, only 36.8% of the TOEFL questions were answered correctly (15.8% when corrected for guessing, using a penalty of 1/3 for each incorrect answer), but using the reduced matrix of rank 300 improves the performance to 64.4% (52.5% corrected for guessing). They claim that the score of 36.8%, using the original matrix, "… is similar to what would be obtained by a mutual information analysis…" (see footnote 5 in [6]).

## 5    TOEFL Experiments

Recall the sample TOEFL question: Given the problem word *levied* and the four alternative words *imposed, believed, requested, correlated*, which of the alternatives is most similar in meaning to the problem word [8]? Table 1 shows in detail how $\text{score}_3$ is calculated for this example. In this case, PMI-IR selects *imposed* as the answer.

Table 2 shows the scores calculated by LSA for the same example [8]. Note that LSA and AltaVista are using quite different document collections for their calculations. AltaVista indexes 350 million web pages [24] (but only a fraction of them are in English). To apply LSA to the TOEFL questions, an encyclopedia was used to create a matrix of 61,000 words by 30,473 articles [8]. However, it is interesting that the two techniques produce identical rankings for this example.

Table 3 shows the results for PMI-IR, for the first three scores, on the 80 TOEFL questions. (The fourth score is not applicable, because there is no context for the questions.) The results for LSA and humans are also presented, for comparison.

**Table 1.** Details of the calculation of $score_3$ for a sample TOEFL question.

| Query | Hits |
|---|---|
| imposed AND NOT (imposed NEAR "not") | 1,147,535 |
| believed AND NOT (believed NEAR "not") | 2,246,982 |
| requested AND NOT (requested NEAR "not") | 7,457,552 |
| correlated AND NOT (correlated NEAR "not") | 296,631 |
| | |
| (levied NEAR imposed) AND NOT ((levied OR imposed) NEAR "not") | 2,299 |
| (levied NEAR believed) AND NOT ((levied OR believed) NEAR "not") | 80 |
| (levied NEAR requested) AND NOT ((levied OR requested) NEAR "not") | 216 |
| (levied NEAR correlated) AND NOT ((levied OR correlated) NEAR "not") | 3 |

| Choice | | $Score_3$ |
|---|---|---|
| p(levied \| imposed) | 2,299 / 1,147,535 | 0.0020034 |
| p(levied \| believed) | 80 / 2,246,982 | 0.0000356 |
| p(levied \| requested) | 216 / 7,457,552 | 0.0000290 |
| p(levied \| correlated) | 3 / 296,631 | 0.0000101 |

**Table 2.** LSA scores for a sample TOEFL question.

| Choice | LSA Score |
|---|---|
| imposed | 0.70 |
| believed | 0.09 |
| requested | 0.05 |
| correlated | -0.03 |

**Table 3.** Results of the TOEFL experiments, including LSA results from [6].

| Interpretation of $p(problem \mid choice_i)$ | Description of Interpretation | Number of Correct Test Answers | Percentage of Correct Answers |
|---|---|---|---|
| $score_1$ | co-occurrence using AND operator | 50/80 | 62.5% |
| $score_2$ | co-occurrence using NEAR | 58/80 | 72.5% |
| $score_3$ | co-occurrence using NEAR and NOT | 59/80 | 73.75% |
| Latent Semantic Analysis | | 51.5/80 | 64.4% |
| Average Non-English US College Applicant | | 51.6/80 | 64.5% |

## 6      ESL Experiments

To validate the performance of PMI-IR on the TOEFL questions, I obtained another set of 50 synonym test questions [5]. Table 4 shows the results of PMI-IR using all four of the different interpretations of p($problem \mid choice_i$).

**Table 4.** Results of the ESL experiments.

| Interpretation of p($problem \mid choice_i$) | Description of Interpretation | Number of Correct Test Answers | Percentage of Correct Answers |
|---|---|---|---|
| $score_1$ | co-occurrence using AND operator | 24/50 | 48% |
| $score_2$ | co-occurrence using NEAR | 31/50 | 62% |
| $score_3$ | co-occurrence using NEAR and NOT | 33/50 | 66% |
| $score_4$ | co-occurrence using NEAR, NOT, and context | 37/50 | 74% |

## 7      Discussion of Results

The results with the TOEFL questions show that PMI-IR (in particular, $score_3$) can score almost 10% higher than LSA. The results with the ESL questions support the view that this performance is not a chance occurrence. However, the interpretation of the results is difficult, due to two factors: (1) PMI-IR is using a much larger data source than LSA. (2) PMI-IR (in the case of all of the scores except for $score_1$) is using a much smaller chunk size than LSA.

PMI-IR was implemented as a simple, short Perl program. One TOEFL question requires eight queries to AltaVista (Table 1).[2] Each query takes about two seconds, for a total of about sixteen seconds per TOEFL question. Almost all of the time is spent on network traffic between the computer that hosts PMI-IR and the computer(s) that host(s) AltaVista. If PMI-IR were multi-threaded, the eight queries could be issued simultaneously, cutting the total time to about two seconds per TOEFL question. If PMI-IR and AltaVista were hosted on the same computer, the time per TOEFL question would likely be a small fraction of a second. Clearly, the hard work here is done by AltaVista, not by the Perl program.

---

[2] For the ESL questions, $score_4$ requires extra queries to select the context word.

The majority of the time required for LSA is the time spent on the SVD. To compress the 61,000 by 30,473 matrix used for the TOEFL questions to a matrix of rank 300 required about three hours of computation on a Unix workstation [6]. A fast SVD algorithm can find a rank $k$ approximation to an $m$ by $n$ matrix $\mathbf{X}$ in time $O(mk^2)$ [25]. Recall that $m$ is the number of words and $n$ is the number of chunks of text. If we suppose that there are about one million English words, then to go from $m \approx 50,000$ to $m \approx 1,000,000$ is an increase by a factor of 20, so it seems possible for SVD to be applied to the same corpus as AltaVista, 350 million web pages [24]. For future work, it would be interesting to see how LSA performs with such a large collection of text.

Several authors have observed that PMI is especially sensitive to the sparse data problem [9]. Landauer and Dumais claim that mutual information analysis would obtain a score of about 37% on the TOEFL questions, given the same source text and chunk size as they used for LSA (footnote 5 in [6]). Although it appears that they have not tested this conjecture, it seems plausible to me. It seems likely that PMI-IR achieves high performance by "brute force", through the sheer size of the corpus of text that is indexed by AltaVista. It would be interesting to test this hypothesis. Although it might be a challenge to scale LSA up to this volume of text, PMI can easily be scaled down to the encyclopedia text that is used by Landauer and Dumais [6]. This is another possibility for future work. Perhaps the strength of LSA is that it can achieve relatively good performance with relatively little text. This is what we would expect from the "smoothing" or "compression" produced by SVD. However, if you have access to huge volumes of data, there is much less need for smoothing.

It is interesting that the TOEFL performance for $score_1$ (62.5%) is approximately the same as the performance for LSA (64.4%) (Table 3). Much of the difference in performance between LSA and PMI-IR comes from using the NEAR operator instead of the AND operator. This suggests that perhaps much of the difference between LSA and PMI-IR is due to the smaller chunk size of PMI-IR (for the scores other than $score_1$). To test this hypothesis, the LSA experiment with TOEFL could be repeated using the same source text (an encyclopedia), but a smaller chunk size. This is another possibility for future work.

Latent Semantic Indexing (LSI) applies LSA to Information Retrieval. The hope is that LSI can improve the performance of IR by, in essence, automatically expanding a query with synonyms [7]. Then a search for (say) *cars* may be able to return a document that contains *automobiles*, but not *cars*. Although there have been some positive results using LSI for IR [8], the results from TREC2 and TREC3 (Text Retrieval Conferences 2 and 3) did not show an advantage to LSI over other leading IR techniques [26]. It has been conjectured that the TREC queries are unusually long and detailed, so there is little room for improvement by LSI [8]. The results reported here for PMI-IR suggest an alternative hypothesis. Most of the TREC systems use a technique called *query expansion* [27]. This technique involves searching with the original query, extracting terms from the top retrieved documents, adding these terms to the original query, and then repeating the search with the new, expanded query. I hypothesize that this query expansion achieves essentially the same effect as LSI, so there is no apparent advantage to LSI when it is compared to an IR system that uses query expansion.

If (say) *cars* and *automobiles* have a high semantic similarity, then we can expect p(*automobiles | cars*) to be relatively high (see equation (2)). Thus, the query *cars* is likely to retrieve a document containing the word *automobiles*. This means that there is a good chance that query expansion will expand the query *cars* to a new query that contains *automobiles*. Testing this hypothesis is another area for future work. The hypothesis implies that LSI will tend to perform better than an IR system without query expansion, but there will be no significant difference between an IR system with LSI and an IR system with query expansion (assuming all other factors are equal).

## 8    Applications

A limitation of PMI-IR is that the network access time for querying a large Web search engine may be prohibitive for certain applications, for those of us who do not have very high-speed, high-priority access to such a search engine. However, it is possible that PMI-IR may achieve good results with a  significantly smaller document collection. One possibility is a hybrid system, which uses a small, local search engine for high-frequency words, but resorts to a large, distant search engine for rare words.

PMI-IR may be suitable as a tool to aid in the construction of lexical databases. It might also be useful for improving IR systems. For example, an IR system with query expansion might use $score_4$ to screen candidate terms for expanding a query. The candidates would be extracted from the top retrieved documents, as with current query expansion techniques. However, current query expansion techniques may suggest sub-optimal  expansion terms, because the top retrieved documents constitute a relatively small, noisy sample. Thus there could be some benefit to validating the suggested expansions using PMI-IR, which would draw on larger sample sizes.

I am particularly interested in applying PMI-IR to automatic keyword extraction [15]. One of the most helpful clues that a word (or phrase) is a keyword in a given document is the frequency of the word. However, authors often use synonyms, in order to avoid boring the reader with repetition. This is courteous for human readers, but it complicates automatic keyword extraction. I am hoping that PMI-IR will help me to cluster synonyms together, so that I can aggregate their frequency counts, resulting in better keyword extraction.

## 9    Conclusions

This paper has introduced a simple unsupervised learning algorithm for recognizing synonyms. The algorithm uses a well-known measure of semantic similarity (PMI). The new contribution is the observation that PMI can exploit the huge document collections that are indexed by modern Web search engines. The algorithm is evaluated using the Test of English as a Foreign Language. The algorithm is  compared with Latent Semantic Analysis, which has also been evaluated using TOEFL. The compari-son sheds new light on LSA, suggesting several new hypotheses that are worth investigating.

## Acknowledgements

Thanks to Joel Martin for many helpful suggestions and general encouragement. Thanks to Eibe Frank, Gordon Paynter, Alain Désilets, Alan Barton, Arnold Smith, and Martin Brooks for their comments on an earlier version of this paper. The 80 TOEFL questions were kindly provided by Thomas K. Landauer, Department of Psychology, University of Colorado.

## References

1. Church, K.W., Hanks, P.: Word Association Norms, Mutual Information and Lexicography. In: Proceedings of the 27th Annual Conference of the Association of Computational Linguistics, (1989) 76-83.
2. Church, K.W., Gale, W., Hanks, P., Hindle, D.: Using Statistics in Lexical Analysis. In: Uri Zernik  (ed.), Lexical Acquisition: Exploiting On-Line Resources to Build a Lexicon. New Jersey: Lawrence Erlbaum  (1991) 115-164.
3. AltaVista, AltaVista Company, Palo Alto, California, http://www.altavista.com/.
4. Test of English as a Foreign Language (TOEFL), Educational Testing Service, Princeton, New Jersey, http://www.ets.org/.
5. Tatsuki, D.:  Basic 2000 Words - Synonym Match 1. In: Interactive JavaScript Quizzes for ESL Students, http://www.aitech.ac.jp/~iteslj/quizzes/js/dt/mc-2000-01syn.html (1998).
6. Landauer, T.K., Dumais, S.T.: A Solution to Plato's Problem: The Latent Semantic Analysis Theory of the Acquisition, Induction, and Representation of Knowledge. Psychological Review, 104 (1997) 211-240.
7. Deerwester, S., Dumais, S.T., Furnas, G.W., Landauer, T.K., Harshman, R.: Indexing by Latent Semantic Analysis. Journal of the American Society for Information Science, 41 (1990) 391-407.
8. Berry, M.W., Dumais, S.T., Letsche, T.A.: Computational Methods for Intelligent Information Access. Proceedings of Supercomputing '95, San Diego, California, (1995).
9. Manning, C.D., Schütze, H.: Foundations of Statistical Natural Language Processing. Cambridge, Massachusetts: MIT Press  (1999).
10. Firth, J.R.: A Synopsis of Linguistic Theory 1930-1955. In Studies in Linguistic Analysis, pp. 1-32. Oxford: Philological Society (1957). Reprinted in F.R. Palmer (ed.), Selected Papers of J.R. Firth 1952-1959, London: Longman (1968).
11. AltaVista: AltaVista Advanced Search Cheat Sheet, AltaVista Company, Palo Alto, California,  http://doc.altavista.com/adv_search/syntax.html (2001).
12. Fellbaum, C. (ed.): WordNet: An Electronic Lexical Database. Cambridge, Massachusetts: MIT Press (1998). For more information: http://www.cogsci.princeton.edu/~wn/.
13. Haase, K.: Interlingual BRICO. IBM Systems Journal, 39 (2000) 589-596. For more information: http://www.framerd.org/brico/.
14. Vossen, P. (ed.): EuroWordNet: A Multilingual Database with Lexical Semantic Networks. Dordrecht, Netherlands: Kluwer (1998). See: http://www.hum.uva.nl/~ewn/.

15. Turney, P.D.: Learning Algorithms for Keyphrase Extraction. Information Retrieval, 2 (2000) 303-336.
16. Grefenstette, G.: Finding Semantic Similarity in Raw Text: The Deese Antonyms. In: R. Goldman, P. Norvig, E. Charniak and B. Gale (eds.), Working Notes of the AAAI Fall Symposium on Probabilistic Approaches to Natural Language. AAAI Press (1992) 61-65.
17. Schütze, H.: Word Space. In: S.J. Hanson, J.D. Cowan, and C.L. Giles (eds.), Advances in Neural Information Processing Systems 5, San Mateo California: Morgan Kaufmann (1993) 895-902.
18. Lin, D.: Automatic Retrieval and Clustering of Similar Words. In: Proceedings of the 17th International Conference on Computational Linguistics and 36th Annual Meeting of the Association for Computational Linguistics, Montreal (1998) 768-773.
19. Richardson, R., Smeaton, A., Murphy, J.: Using WordNet as a Knowledge Base for Measuring Semantic Similarity between Words. In Proceedings of AICS Conference. Trinity College, Dublin (1994).
20. Lee, J.H., Kim, M.H., Lee, Y.J.: Information Retrieval Based on Conceptual Distance in IS-A Hierarchies. Journal of Documentation, 49 (1993) 188-207.
21. Resnik, P.: Semantic Similarity in a Taxonomy: An Information-Based Measure and its Application to Problems of Ambiguity in Natural Language. Journal of Artificial Intelligence Research, 11 (1998) 95-130.
22. Jiang, J., Conrath, D.: Semantic Similarity Based on Corpus Statistics and Lexical Taxonomy. In: Proceedings of the 10th International Conference on Research on Computational Linguistics, Taiwan, (1997).
23. Brin, S., Motwani, R., Ullman, J., Tsur, S.: Dynamic Itemset Counting and Implication Rules for Market Basket Data. In: Proceedings of the 1997 ACM-SIGMOD International Conference on the Management of Data (1997) 255-264.
24. Sullivan, D.: Search Engine Sizes. SearchEngineWatch.com, internet.com Corporation, Darien, Connecticut, http://searchenginewatch.com/reports/sizes.html (2000).
25. Papadimitriou, C.H., Raghavan, P., Tamaki, H., Vempala, S.: Latent Semantic Indexing: A Probabilistic Analysis. In: Proceedings of the Seventeenth ACM-SIGACT-SIGMOD-SIGART Symposium on Principles of Database Systems, Seattle, Washington (1998) 159-168.
26. Sparck Jones, K.: Comparison Between TREC2 and TREC3. In: D. Harman (ed.), The Third Text REtrieval Conference (TREC3), National Institute of Standards and Technology Special Publication 500-226, Gaithersburg, Maryland (1994) C1-C4.
27. Buckley, C., Salton, G., Allan, J., Singhal, A.: Automatic Query Expansion Using SMART: TREC 3. In: The Third Text REtrieval Conference (TREC3), D. Harman (ed.), National Institute of Standards and Technology Special Publication 500-226, Gaithersburg, Maryland (1994) 69-80.